\def\year{2020}\relax
\documentclass[letterpaper]{article} 
\usepackage{aaai20}  
\usepackage{times}  
\usepackage{helvet} 
\usepackage{courier}  
\usepackage[hyphens]{url}  
\usepackage{graphicx} 
\usepackage{graphicx}  
\title{Self-supervised reinforcement learning for  speaker localisation with  the iCub humanoid robot }
\author{Jonas Gonzalez-Billandon\textsuperscript{\rm 1,2}, Lukas Grasse\textsuperscript{\rm 3}, Matthew Tata\textsuperscript{\rm 3}, Alessandra Sciutti\textsuperscript{\rm 1},\\ 
 \Large{\textbf{Francesco Rea\textsuperscript{\rm 4}}} \\ 
\textsuperscript{\rm 1}Italian Institute of technology, CONTACT\\ 
\textsuperscript{\rm 2}University of Genova, DIBRIS\\
\textsuperscript{\rm 3}University of Lethbridge\\
\textsuperscript{\rm 4}Italian Institute of technology, RBCS\\

Center for Human Technologies Via Enrico Melen 83\\
Italy, Genova,  16152\\
jonas.gonzalez@iit.it 
}
 
\begin{document}

\maketitle

\begin{abstract}
In the future robots will interact more and more with humans and will have to communicate naturally and efficiently. Automatic speech recognition systems (ASR) will play an important role in creating natural interactions and making robots better companions. Humans excel in speech recognition  in noisy environments and are able to filter out noise. Looking at a person's face is one of the mechanisms that humans rely on when it comes to filtering speech in such noisy environments. Having a robot that can look toward a speaker could benefit ASR performance in challenging environments. To this aims, we propose a self-supervised reinforcement learning based framework inspired by the early development of humans to allow the robot to autonomously create a dataset that is later used to learn  to localize speakers with a deep learning network. 
\end{abstract}


\section{Introduction}
Speech recognition systems often do not match the published benchmark performance of datasets when applied to robotics platforms \cite{Sunderhauf2018}. This drop in performance can be explained by the variability and noise of auditory environments which are difficult to capture in a representative dataset. Dealing with these noisy and unstructured environments is a challenging and well-known problem in the speech community known as the "cocktail party problem"  \cite{haykin2005cocktail}. The human brain excels at this task and is able to isolate and focus on a speech source and maintain conversations in these noisy environments. Research has shown that viewing a speaker’s face enhances speech recognition in noisy environments \cite{Golumbic}. Localizing a target speaker and being able to orient towards him is, therefore, an important skill that could also benefit ASR in robotics. In this work, we address this point following a self-supervised  learning framework. We propose a computational framework that enables the iCub humanoid robot to learn to localize and recognize speakers using cross-modal input and reinforcement learning. We take inspiration from the human brain and use phase-locking across the auditory and visual modalities to identify the active speaker. We use reinforcement learning to guide the robot to find the active speaker in the scene which we use to create a dataset of audio inputs and their spatial location. The reinforcement learning is used to create a dataset that is later used to train a deep learning network for speaker localisation. The goal of our framework is to propose :
\textbf{(i)} a mechanism to focus on an active speaker by learning an optimal motor policy using visual and auditory signal and \textbf{(ii)} use the ability to locate an active speaker to develop a faster and more accurate speaker localisation mechanism working only with an auditory signal.


\begin{figure*}[!th]
    \includegraphics[width=18cm, height=3.5cm]{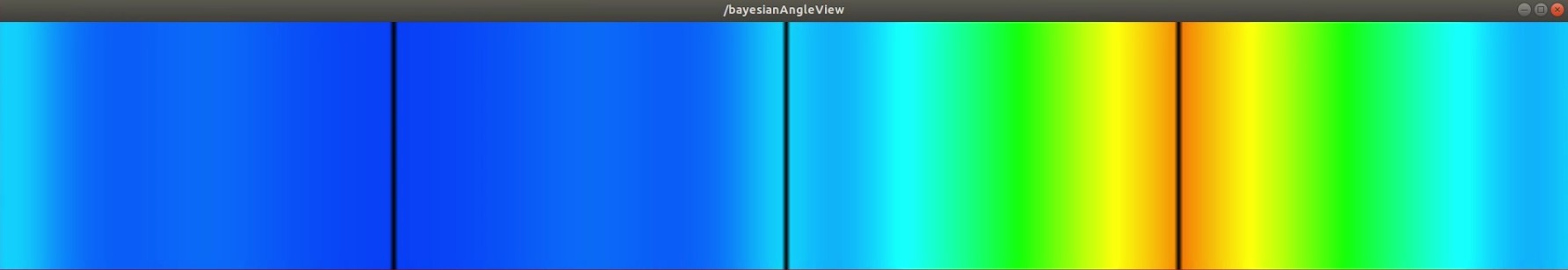}
    \caption{ Attention bayesian map of possible sound location: The red bar indicates the maximum probability for the sound source location}
\end{figure*}

\section{Background}
Localising sound is a crucial skill that helps humans understand others and the environment \cite{Ebata}. Newborns can orient towards a sound source in their first months of life but their system is limited and slow.  Muir and Field show that newborns sound localisation is slow and can only discriminate vague positions. They demonstrate that babies needs continuous sound exposure with long durations in order to discriminate between left and right location \cite{Muir1979}. After few months (5-6) babies are much faster in discriminating between left and right sound sources. How does the human sound localisation system calibrate itself? How do infants know to associate a sound with a spatial location and perfect their skills?\\

Experience plays an important role in the calibration of the sound localisation system. Multi-sensory coordination and more specifically audio visual matching help infants to develop their sound localisation system. The pairing of visual stimuli with a certain pattern of sound leads to the calibration of the sensory space \cite{Gori2012}. For example, it is well known that babies show preference for human voices over any other sound \cite{vouloumanos2007listening} but also for human faces \cite{Farroni2005}. This results in a baby being more attentive to voice and faces giving a facilitation mechanism to associate the voice with a face. Exposure to people allow babies thanks to  their attentional mechanism to calibrate their sound localisation system by matching speech with faces. This matching arises from a temporal matching between lips movements and speech frequency \cite{dodd1979lip}. This particular skill allows the autonomous calibration of their sound localisation system driven by their own experiences.


\section{Implementation}
 Our objective is to train a  speaker localisation system in a self-supervised manner inspired by the early development of humans. Similar to how we gradually improve our  ability to localize sounds in the environment we want to replicate this mechanism with the iCub robot. To do so we use a reinforcement learning (RL) framework to learn the optimal motor policy to focus on an active speaker using visual and auditory signals (face position, speaker vague position). The learned policy allows us to  create a suitable dataset $D = (x,Y)$ where $x$ is binaural audio inputs and  $Y$ is the egocentric position in the robot reference frame  defined by the two angles: $Azimuth$ and $Elevation$. The dataset is then used to learn a faster and more accurate sound localisation model by training a deep neural network. The  advantage is that the data needed to train the sound localisation system are autonomously created with the RL algorithm. Moreover, when the trained speaker localisation system achieves good performances it can bypass the RL framework using only auditory signals to localise the active speaker as it can happen in humans.


\subsection{Early sound source localisation - Audio attention}
To reproduce the early auditory attentional mechanism of newborns \cite{wertheimer1961psychomotor} we use a biologically inspired audio attention system. The  localisation is build by approximating the spectral decomposition of the human basilar membrane with a Gammatone filterbank and modelling delay-tuned units in the auditory pathway as banks of narrow-band delay-and-sum beamformers using the interaural time difference (ITD). The auditory salience is then used in a Bayesian framework to build a distribution of probabilities of sound source locations across a map of azimuthal angles \textbf{Fig 1} \cite{Hambrook}. This system provides evidence about the instantaneous sound sources in the auditory scene. The system aimed at reproducing the early behaviours of babies with a limited resolution in space and relatively long exposure to perform well.\\

We use a fuzzy logic membership approach to decompose the auditory scene into five spatial locations (\textbf{Fig 2}) \cite{zadeh1975concept}. The estimated sound location $a$ from the Bayesian attentional module in the robot’s allocentric coordinate frame is mapped to the linguistic variable $T(L) \in $ $\{far\_left, left, center, right, far\_right\}$ with  a fuzzy membership $\mu_t(a)$. 
The final location $L$ is then computed following equation 1.
\begin{equation}
    L = max(\mu_t(a))  \hspace{0.2cm} \forall \mu_t \in T(L)
\end{equation}

\begin{figure}[!th]
    \centering
    \includegraphics[width=9cm, height=4cm]{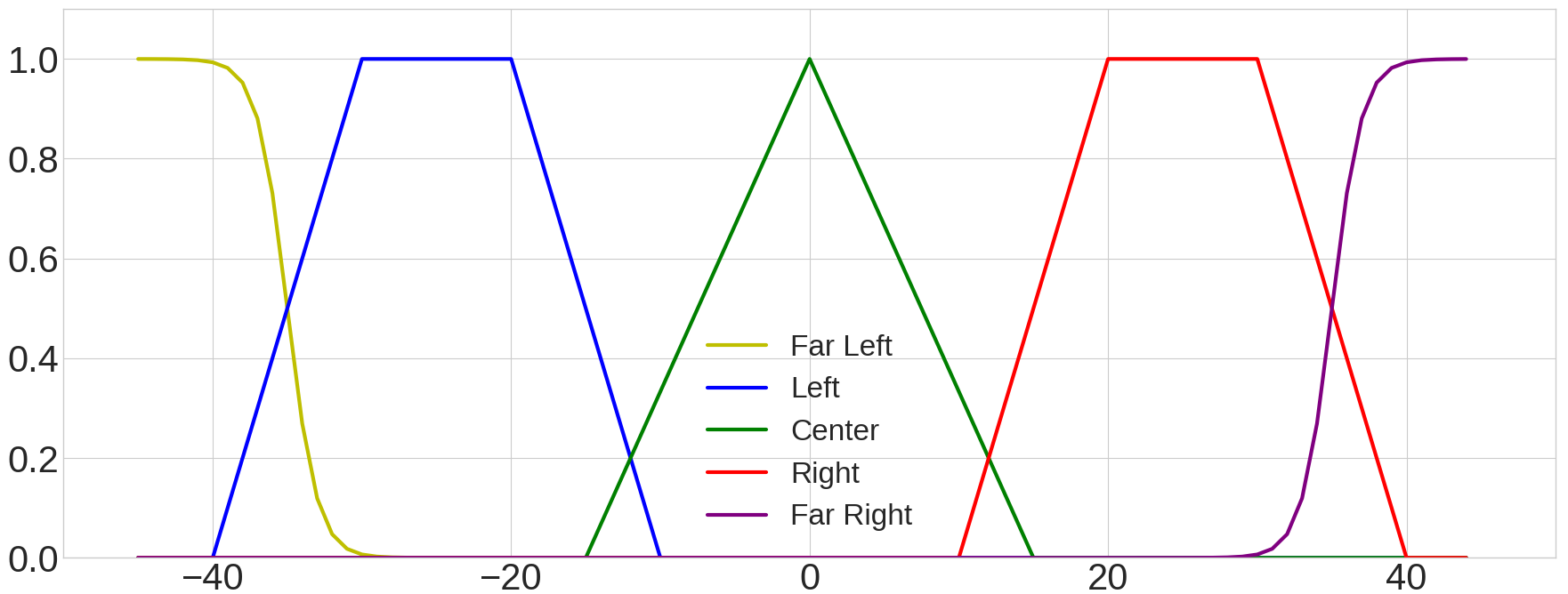}
    \caption{ Fuzzy membership of $T(L)$ define by the head angles range.}
\end{figure}

\subsection{Identification of an active speaker}
 Given the vague estimation of $a$ we want to learn the motor policy $\pi$ to successfully move the robot's head toward the active speaker. To achieve this we want to use a reinforcement learning framework defined by the triplet: $(S, A, R)$.
 \newline
 \noindent
 
 \subsubsection{\textbf{State S}}
 \boldmath
 The state $S$ will have features as the combination of the source speech localisation represented by $L$, the robot head position $\Theta$ and a bio-inspired face detection system. The linguistic variable $L$ will be transformed to a one hot vector $\textbf{V}^{ 1 \times N}$ where $N$ is the length of the set $T(L)$. The face localisation system output will be represented as a matrix $\textbf{M}^{w \times h} \in [0,1]$  where $w$ and $h$ defined the visual field of view of the robot and a 1 indicates the presence of a face and zero no face present. Finally the vector $\Theta$ will represent the state of the robot head defined by two values ($\Theta^1, \Theta^2$) for the pan and tilt angles. So the state will be defined as $S_t=\{O_1,...,O_t\}$ where $O_t =(V_t,M_t, \Theta_t)$ and $t$ defined the time window. 
\newline

\noindent
\subsubsection{\textbf{Action $A$}}
 The action $A$ that the robot will have to learn will be the two degrees of freedom (pan and tilt) needed to look at the speaker. The set of actions will be $A = \{  \emptyset , \leftarrow, \uparrow,  \downarrow, \rightarrow \}$ for doing nothing turning the head left, upward, downward and right. 
\newline

\noindent
\subsubsection{\textbf{Reward R - Phase-locked correlation}}
We propose to use as  reward function $R$ a combination of two rewards $R_{face}$ and $R_{corr}$. $R_{face}$ represents if the robot is fixating a face while $R_{corr}$ is a value computed from the  correlation of the speech envelope  $E_{speech}$ with $M_{area}$. $M_{area}$ is defined as the inter  lips area extracted from the face localisation system and a mouth landmarks detection algorithm.

\begin{equation}
        R = R_{face} + R_{corr}
\end{equation}

Following this reward function, the reward signal will be maximal when facing a face with the matching voice. $R_{face}$ will constraint the learning to look at faces which enable the extraction of the mouth area.


\section{Self-supervised learning } 
In self-supervised learning the training data is automatically but approximately labeled by finding and exploiting the relations or correlations between inputs. In our case we want to associate the spatial location $Y$ with the corresponding audio input $x$. The idea is to use the interaction between vision and audio to provide the label $Y$.  When the audio attention module identifies a source speech the binaural audio can be saved, creating evidences $x$ but without an associated label $Y$. By using the learned motor policy $\pi$ to focus on the active speaker we can match the previous evidence $x$ with the active speaker. A label $Y$ can then be extracted using the proprioceptive information of the robot head (pan and tilt). This will result in the automatic creation of a dataset in an unsupervised way which will be used to train a classical supervised learning system such as deep learning network. The advantage of using a model for sound localisation rather than relying only on the learned policy $\pi$ is the possibility to have a faster and more accurate system working with only one modality.



\section{Methodology}
We focus on testing the correlation between the extracted mouth area and the speech envelope from the robot sensors. This is the most critical part of our framework as it serve as a reward signal to guide the robot to focus on the active speaker and consequently build the dataset autonomously. We asked 5 participants (3 males, 2 females) with four different languages (see \textbf{Table I}) to read a story in front of the robot at different distances. Participants read the story for an average duration of 3 minutes with a range of distance of 50 centimeters to 3 meters from the robot. Then to further test the validity of this approach to identify the active speaker we asked two people to speak at turn in front of the robot as illustrated in \textbf{Fig 3}.  

\subsection{Face and mouth area extraction}
The iCub robot has two cameras that can stream images  $I_l$, $I_r$ of size 720x570 . We use only the  image $I_l$ to detect the face and extract the mouth landmarks. We use use a bio-inspired face detection system \cite{Gonzalez-billandon2020} to detect the face. Once a face was detected we extracted the mouth landmarks with the  DLIB facial landmark detector (see \textbf{Fig 3}). From the mouth landmarks we computed the mouth area $M_{area}$ on real time achieving  a sampling rate $M_{sr}$ of 10 points by seconds. 

\begin{figure}[!th]
    \centering
    \includegraphics[scale=0.3]{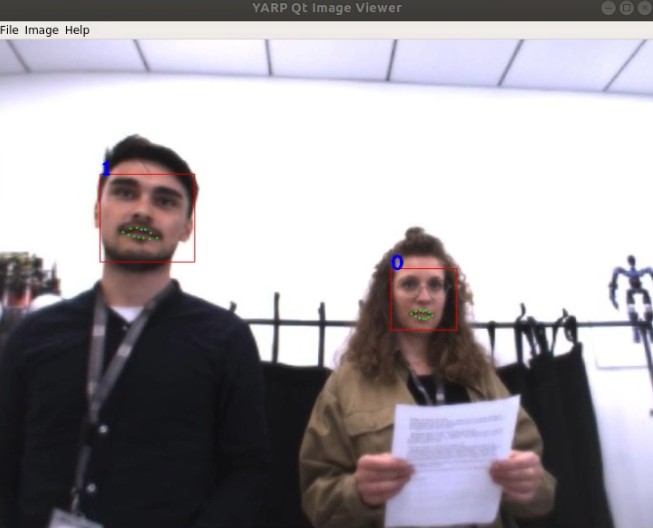}
    \caption{ Example of detection of faces with mouth landmarks}
\end{figure}

\subsection{Speech envelope}
The iCub robot has two microphone that stream a 2-channel audio signal $S$ at 48Khz. We compute the amplitude envelope by taking the magnitude of the analytic signal $S$. We use the Hilbert transform to determine the amplitude envelope on both channels resulting in two envelopes $E_1$, $E_2$. We re-sample the two envelopes to the sampling rate of the mouth area signal $M_{area}$ to perform the correlation.

\subsection{Pearson correlation}
Given the two envelopes $E_1$, $E_2$ and the $M_{area}$   we compute the Pearson correlation on a time window $t$. We then take the $r$ value with the minimum $p$ value from the two channels.


\section{Results}
We run the Pearson correlation with $t$ set at 10 seconds resulting with a $M_{sr}$ of 100 data points per window. Participants talked for 2 minutes on average leading to 12 different windows. We perform the correlation on all the windows and reported the average $r$ values and the percentage of significant $p$ values (\textbf{Table I}).\\

\noindent
We examine the correlation of two people speaking at turn in front of the robot to highlight the change of correlation value through time \textbf{Fig 5}. 

\begin{figure*}[!th]
    \includegraphics[width=18cm, height=5.5cm]{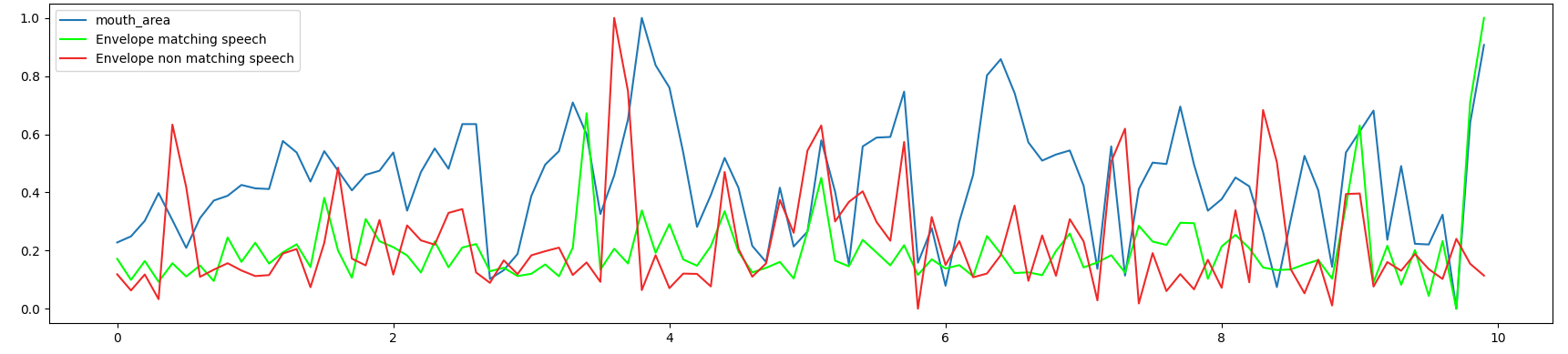}
    \caption{Pearson correlation for the mouth area  function with  the matching speech envelope ($r=0.48$, $p < 1e^{-5}$) and non matching speech envelope ($r=0.01$, $p > 0.05$) with a time window of 10 seconds.}
\end{figure*}

\begin{figure}[!th]
    \leftskip -0.3cm
    \includegraphics[scale=0.3]{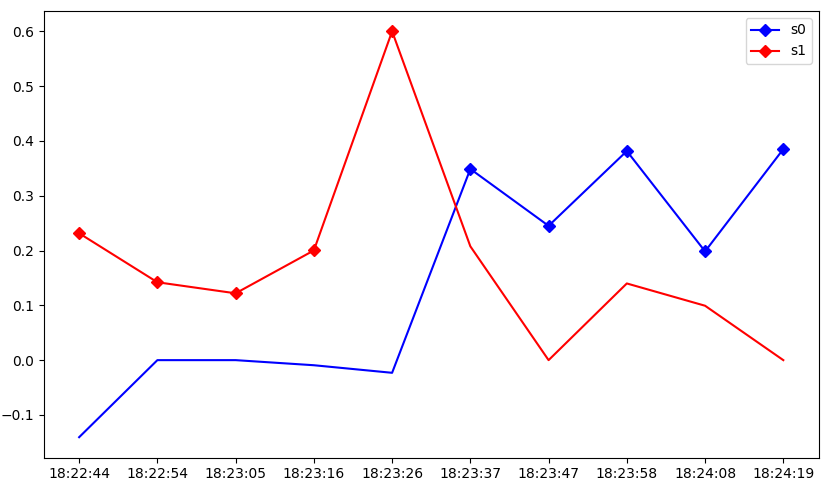}
    \caption{ Correlation values on 10 windows of 10 seconds for two speakers, diamond shape represent when $p \textless 0.05$.}
\end{figure}


\begin{table}[!h]
\centering
\begin{tabular}{|c|c|c|c|} 
\hline
Participants & Languages & Mean correlation &  p \textless 0.05   \\ 
\hline
s0           & French    &  $0.40 \pm 0.14$ & 80\%  \\ 
\hline
s1           & Italian   & $0.47 \pm 0.15$  & 89\%  \\ 
\hline
s2           & Italian   & $0.40 \pm 0.24$ & 80\%   \\ 
\hline
s3           & Spanish   & $0.37 \pm 0.17$ & 63\%   \\ 
\hline
s4           & English   & $0.52 \pm 0.13$  & 88\%   \\
\hline
\end{tabular}
 \caption{Average correlation of mouth area with speech envelope for each subjects for  2 minutes recording}
\end{table}

\section{Discussion}
In this work we want to take advantage of the embodiment of the robot to re-propose the optimal strategy used by the human brain to create a self-supervised learning mechanism that learns to localise speakers. We presented the different parts of the framework and tested the critical part of the $R_{corr}$ function that will guide the robot to recognize the active speaker. $R_{corr}$ is defined as the correlation between the inter-lips area and the speech envelope. Results show that we can successfully extract both information in real-time from the robots sensors. In one to one interaction with the robot the correlation values are positive and statistically significant and correlated only with the true speaker \textbf{Fig 4}.  Furthermore, when tested in a more realistic scenario where two people speak at turn in front of the robot the correlation values can be use to discriminate the active speaker \textbf{Fig 5}. Together these results validate our approach to use mouth area and speech envelope value as a reward signal to guide the robot to focus on the active speaker. We focused on testing the most critical part of our framework the $R_{corr}$ function as the other parts of our framework are modules that we have already developed and tested on the robot. After validating that the correlation between the inter-lips area and the speech envelope can be used to recognize an active speaker the next step will be to run the reinforcement learning. By using fuzzy logic to represent the sound source localisation the state space is constrained, which will facilitate the learning. We will adopt a transfer learning approach where first we will create a virtual environment to train and then do domain adaptation on live interactions with the robot. We will also focus on optimizing the mouth area extraction to increase the sampling rate which will allow to reduce the time window and have a more reactive framework.

\section{Conclusion}
We presented a bio-inspired framework to allow the iCub robot to learn to localise speaker in the environment in an autonomous way. Our framework  aims at reproducing the multi-modal  learning of babies, and more precisely cross-sensory calibration. We took inspiration on how babies calibrate their auditory space using vision and motor actions and apply it to the iCub robot. We proposed a self-supervised learning approach for the task of speaker localisation using auditory and visual inputs. Starting with an early localisation system as in babies we used a reinforcement learning algorithm to solve the multi-sensory integration problem  to recognise an active speaker in the scene. This allow us to create a dataset of audio, location mapping which can be use to train  a more robust and accurate speaker localisation system.  We gave details of the different parts of our framework and tested the critical part of our solution: the speech mouth correlation. We demonstrated that we can recognize in real-time an active speaker using speech-mouth correlation. Future work, will be to implements the other parts of the framework and test it with the iCub robot on live interactions.

\bibliographystyle{aaai}
\bibliography{manuscript}

\end{document}